\documentclass[conference]{IEEEtran}
\IEEEoverridecommandlockouts
\usepackage{cite}
\usepackage{amsmath,amssymb,amsfonts}
\usepackage{algorithmic}
\usepackage{graphicx}
\usepackage{textcomp}
\usepackage{xcolor}
\usepackage{url}
\usepackage{soul}

\def\BibTeX{{\rm B\kern-.05em{\sc i\kern-.025em b}\kern-.08em
    T\kern-.1667em\lower.7ex\hbox{E}\kern-.125emX}}
\begin{document}

\title{Toward the Automated Construction of Probabilistic Knowledge Graphs for the Maritime Domain}
\author{\IEEEauthorblockN
{{\small 1\textsuperscript{st}} Fatemeh Shiri, 
{\small 4\textsuperscript{th}} Teresa Wang\\ 
{\small 5\textsuperscript{th}} Shirui Pan,  
{\small 6\textsuperscript{th}} Xiaojun Chang\\
{\small7\textsuperscript{th}} Yuan-Fang Li,  
{\small 8\textsuperscript{th}} Reza Haffari} 
\IEEEauthorblockA{\textit{Department of Data Science and AI} \\
\textit{Faculty of Information Technology} \\
\textit{Monash University}\\
Melbourne, Australia \\
firstname.lastname@monash.edu\vspace{-20pt}\vspace{-5pt}}
\and
\IEEEauthorblockN{ {\small 2\textsuperscript{nd}} Van Nguyen}
 \IEEEauthorblockA{\textit{Intelligence, Surveillance and Space Division} \\
\textit{Defence Science and Technology Group}\\
Adelaide, Australia \\
Van.Nguyen5@dst.defence.gov.au}
\and
\IEEEauthorblockN{{\small 3\textsuperscript{rd}} Shuang Yu}
\IEEEauthorblockA{
\textit{Monash University}\textsuperscript{*}\\
Melbourne, Australia \\
Shuang.Yu@ibm.com}
\textsuperscript{*}\tt\textsuperscript{ now with IBM Australia}}

\def\FS#1{{\color{blue}{#1}}}
\def\YF#1{{\color{purple}{#1}}}
\def\VN#1{{\color{red}{#1}}}
\def\XC#1{{\color{green}{#1}}}
\def\RH#1{{\color{magenta}{#1}}}
\def\TW#1{{\color{brown}{#1}}}
\def\SP#1{{\color{cyan}{#1}}}
\def\SY#1{{\color{teal}{#1}}}

\maketitle

\begin{abstract}

International maritime crime is becoming increasingly sophisticated, often associated with wider criminal networks. Detecting maritime threats by means of fusing data purely related to physical movement (i.e., those generated by physical sensors, or \textit{hard} data) is not sufficient. This has led to research and development efforts aimed at combining hard data with other types of data (especially human-generated or \textit{soft} data). Existing work often assumes that input soft data is available in a structured format, or is focused on extracting certain relevant entities or concepts to accompany or annotate hard data. Much less attention has been given to extracting the rich knowledge about the situations of interest implicitly embedded in the large amount of soft data existing in unstructured formats (such as intelligence reports and news articles). In order to exploit the potentially useful and rich information from such sources, it is necessary to extract not only the relevant entities and concepts, but also their semantic relations, together with the uncertainty associated with the extracted knowledge (i.e., in the form of probabilistic knowledge graphs). This will increase the accuracy of, and confidence in, the extracted knowledge and facilitate subsequent reasoning and learning. To this end, we propose Maritime DeepDive, an initial prototype for the automated construction of probabilistic knowledge graphs from natural language data for the maritime domain. In this paper, we report on the current implementation of Maritime DeepDive, together with preliminary results on extracting probabilistic events from maritime piracy incidents. This pipeline was evaluated on a manually crafted gold standard, yielding promising results.
\end{abstract}

\vspace{-2.2mm}
\begin{figure*}[htbp]
\scalebox{1}[0.95]{\includegraphics[width=1\linewidth]{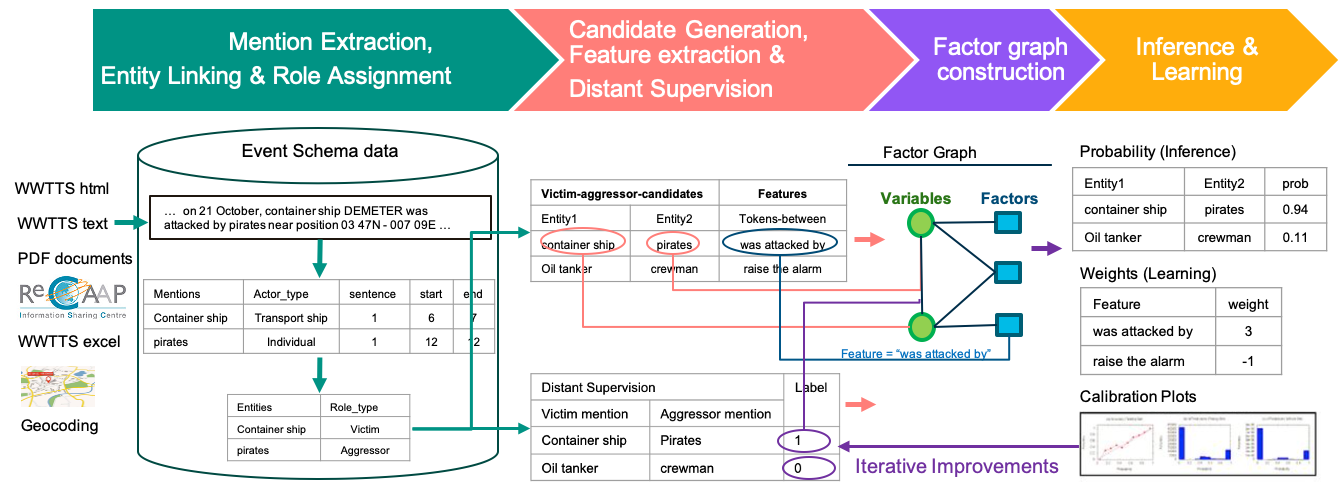}}
\vspace{-0.5em}
\caption{Overview of the proposed framework, Maritime DeepDive.}
\label{fig:pipeline}
\vspace{-1.4em}
\end{figure*}

\section{Introduction and Background}
\vspace{-1.2mm}
Maritime security has increasingly been recognised as a major global security concern. Due to the increased shipping traffic (i.e., 80$\%$ of all world trade is transported by sea, and seaborn trade has doubled every decade since 1945 \cite{rosenberg2009political}) and lack of agreed jurisdiction for maritime enforcement, maritime security threats such as piracy have been on the rise. This poses a significant threat to the economy (e.g., piracy worldwide has cost an average of tens of billion US dollars a year), to humanitarianism (e.g., 3,862 seafarers were attacked and 561 hostages captured in Somali in 2011), and to our environment (e.g., pollution damage caused by oil and chemical spills) \cite{hurlburt2013human}.

Due to the vast territorial waters involved, patrolling using military resources (such as the EU's Operation Atlanta\footnote{Operation ATALANTA. \url{https://eunavfor.eu/}.}, NATO’s Operation Ocean Shield\footnote{Operation OCEAN SHIELD. \url{https://mc.nato.int/missions/operation-oceanshield}.} and the US-led Combined Task Force-151\footnote{CTF 151: Counter Piracy. \url{https://combinedmaritimeforces.com/ctf-151-counter-piracy/}.}) is only partially successful \cite{dabrowski2014contextual}. In addition, organised maritime crime is often associated with wider criminal networks, e.g., there are connections between organised piracy and wider criminal networks and corruption on land \cite{rogova2013contextual}. As such, many associated criminal activities can be partially or completely carried out through natural language communications (e.g., information manipulation or attack planning) \cite{rein2018beyond}, and thus may not be detectable by purely physical movement data provided by physical sensors.

Much research and development has been directed toward the automated processing, fusion and reasoning with diverse data sources for maritime surveillance to assist human analysts with decision making. In order to make sense of complex scenarios to detect maritime threats, and for situational awareness in general, it has been increasingly recognised that the fusion of data relating to physical movement, such as that generated by physical sensors (i.e., hard data, which has been extensively studied for sensor data fusion), is not sufficient, but requires all available information, including data provided by humans and intelligence (i.e., soft data) as well as other operational and contextual information \cite{rein2018beyond, dragos2014integration}. Efforts to incorporate soft data into fusion frameworks for situational awareness have considered various types of soft data, including extant textual data in a structured or logical format \cite{wickramarathne2011belief, nunez2013hard}; text annotations associated with hard data that potentially provide additional attributes for the relevant entities \cite{dragos2015critical}; certain entities and concepts of interest \cite{plachkov2016soft,wu2014semantic, panchapakesan2019optimizing,Abirami2015,burks2019collaborative} and a few simple relations between them \cite{Buffett2017arctic}  extracted from short human statements \cite{panchapakesan2019optimizing}, social network data \cite{Abirami2015}, semi-structured synthesised reports \cite{burks2019collaborative}, maritime incident reports \cite{plachkov2016soft} and news articles \cite{Buffett2017arctic}; and also semantic knowledge constructed from natural language text (albeit, without taking into account the \emph{uncertainty} associated with the extracted knowledge) ~\cite{shapiro2015use,zhang2020construction}.

To this end, Reasoning under Uncertainty with Soft and Hard data (RUSH) \cite{nguyen2019combining,nguyen2020fuzzy} was proposed as a framework for reasoning and learning from diverse sources for real-world situational awareness. With respect to soft data in particular, RUSH is motivated by harnessing human knowledge implicitly embedded in abundant natural language data often existing in many real-world scenarios (e.g., news articles and intelligence reports). 
This form of complex soft data is unstructured and potentially contains substantial knowledge about a (maritime) situation.
This includes certain entities or spatial/temporal concepts of interest, and expressing the relationships between them with a clear and coherent semantics. This facilitates subsequent reasoning and learning while managing the associated uncertainty (crucial for handling noisy data such as natural language), and enables explanation and provenance (essential for informing decision-making).
In other words, we are interested in constructing \emph{probabilistic knowledge graphs}, where each piece of extracted knowledge is associated with a marginal probability. This paper discusses our initial effort toward achieving this goal, that is, extracting probabilistic knowledge for maritime events from piracy incident reports.

\section{Knowledge graph construction}
\vspace{-1.2mm}

Knowledge graphs are used by many information systems that require access to structured knowledge. Currently, the most popular knowledge graphs in the research context (e.g. Google \cite{pelikanova2014google}, Freebase \cite{bollacker2008freebase}, YAGO \cite{rebele2016yago}, and DBpedia \cite{lehmann2015dbpedia}) are designed to be generic. These general-purpose knowledge graphs consist of several million entities and relations. They are typically wide-breadth, large-scale, and exhibit a high level of automation and openness. Despite these favourable characteristics, their application to specific domains is limited due to a lack of domain-specific knowledge, resulting in a low degree of accuracy and consequently an inability to reason effectively. 

Due to the limitations of generic knowledge graphs, domain-specific knowledge graphs are necessary for critical applications. Such knowledge graphs are based on industry-specific ontologies and data, and generally achieve higher accuracy in their target domains. They are typically designed for knowledge management, data governance and aggregation of industry information, and assist with various AI and machine learning tasks. For instance, in the field of medical sciences, well-known domain-specific knowledge graphs include ``Knowlife'' \cite{ernst2015knowlife} and a knowledge graph of disease symptoms \cite{rotmensch2017learning}. There have also been knowledge graphs constructed in other industries, such as an academic knowledge graph\mbox{\cite{farber2019microsoft}}, and the knowledge graph on education ``KnowEdu'' \cite{chen2018knowedu}.

In respect to the maritime domain specifically, there are a few studies on knowledge graphs and their construction \cite{zhang2019construction, everwyn2019link}; for instance \cite{zhang2019construction} is concerned with knowledge graphs for maritime dangerous goods, and for the prediction of vessel locations. However, these efforts are either aimed at constructing knowledge graphs from hard (i.e., AIS) data (as in \cite{everwyn2019link}); or focused on extracting information from soft data in a structured format (e.g., the \textit{International Maritime Dangerous Goods Code}) rather than from unstructured data such as textual reports (as in \cite{zhang2019construction}). 
In addition, due to the need for domain-specific data and the strong dependence on experts in the industry, the construction of comprehensive domain-specific knowledge graphs is not straightforward and can be very time consuming. To this end, we propose Maritime DeepDive, a framework that facilitates the construction of a (probabilistic) knowledge graph from natural language data. Maritime DeepDive improves on existing work in that the system extracts not only maritime entities of interest and their attributes, but also their semantic relations, dealing with uncertainty and computing the degree of confidence associated with the extracted knowledge. We envisage that Maritime Deepdive will be able to assist human experts by automating many tasks through machine-human interactions. 


\section{Maritime DeepDive}
\vspace{-1.2mm}
In this section we describe our approach, Maritime DeepDive, and provide a demonstration of events and activities of interest extracted from maritime piracy incident reports.

To build a comprehensive knowledge graph, we employ the publicly available platform DeepDive \cite{de2016deepdive} which was developed for efficient and interpretable machine learning and statistical inference to facilitate explainability and provide provenance. DeepDive can effectively address the problems of extraction, cleaning, and integration jointly, and treat them as a single probabilistic inference problem that takes all available information into account. 
Specifically, DeepDive was formulated on top of a large-scale inference engine based on Markov Logic \cite{niu2011tuffy}, currently used as part of the high-level fusion and reasoning capability within RUSH for real--world situational awareness \cite{niu2011tuffy, nguyen2020fuzzy}. This in effect allows Maritime DeepDive to (i) model the complexity and uncertainty of the extracted knowledge from natural language, (ii) combine learning from both expert knowledge and available data (via distant supervision), and (iii) facilitate future combination with probabilistic knowledge and events generated from hard data within RUSH \cite{nguyen2020fuzzy} in a consistent and coherent manner. Fig~\ref{fig:pipeline} demonstrates the architecture of our end-to-end framework. 


To build the maritime knowledge graph in Maritime DeepDive, we carry out a few data preprocessing steps. The information from different sources is collated and preprocessed to ensure integrity, and then merged. The next step is to add Natural Language Processing (NLP) markups and extract candidate relation mentions together with a sparse feature representation for each candidate. The extracted candidate relation mentions include mentions associated with various entities of interest, their roles and potential relationships between candidates. This information is used in the heuristic rules we developed to perform distant supervision. Finally, a high-level configuration of our model is specified, so the maritime probabilistic knowledge graph can be (automatically) constructed. The following subsections provide details of the components of this architecture and discuss how they are applied to build a knowledge graph for the maritime domain.

\begin{figure*}[htbp]
\scalebox{1}[0.90]{\includegraphics[width=1\linewidth]{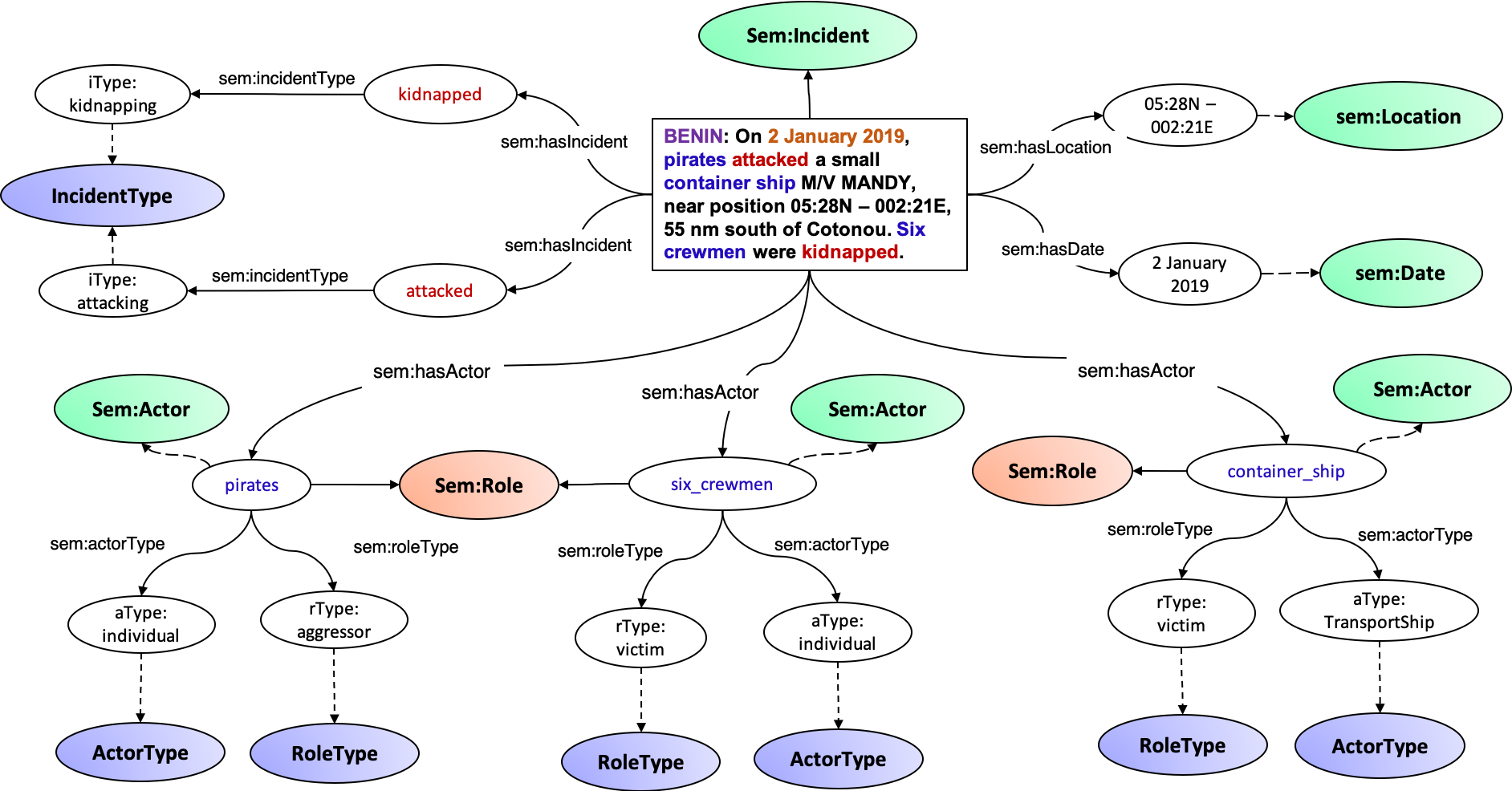}}
\caption{An example graph of a piracy report modeled in SEM including role types}
\label{fig:SEM}
\vspace{-1.1em}
\end{figure*}

\subsection{Data Sources}
\vspace{-1.2mm}
\label{data}
We obtained relevant data from two main sources: (a) the Worldwide Threats To Shipping (WWTTS)\footnote{\url{https://msi.nga.mil/Piracy}} and (b) the Regional Cooperation Agreement on Combating Piracy and Armed Robbery against Ships in Asia (ReCAAP)\footnote{\url{https://www.recaap.org/}}. The WWTTS reports are obtained from the Office of Naval Intelligence, which compiles and publishes the reports on a weekly basis. They contain summaries of piracy and hostility acts towards worldwide commercial shipping since 1994. The content in these reports is unstructured data and/or semi-structured or structured data (e.g.\ HTML, XML or CSV files). The ReCAAP documents report piracy and armed robbery against ships weekly from 2006 up to the present. The reports contain several types of criminal activities, including \emph{attacking}, \emph{boarding}, \emph{firing upon}, \emph{kidnapping}, \emph{hijacking}, \emph{robbery} and \emph{suspiciously approaching}. 

These data repositories have been widely used as sources of soft/natural language data to assist with the detection and analysis of maritime piracy events and behaviours. However, unlike the majority of the existing work where the structured data (i.e., in the form of a database) provided by the repositories are used as the input, the input data in our work includes the textual incident reports. We use the structured data (together with domain and expert knowledge) as sources of distant supervision for training the probabilistic labels, from which knowledge and events related to maritime piracy can be extracted from the unstructured data.

We extracted all incidents from the WWTTS and ReCAAP documents for the most recent five years (2015--2020) where the incident includes a paragraph describing piracy activity. We sorted all these incidents, removed duplicates and maintained them in a relational database (Postgres). In total, we obtained 1,940 incidents. We randomly selected 75 incidents as the test dataset and the rest (1865 incidents) were used for training. 
For each incident we applied a markup produced by standard NLP preprocessing tools\footnote{\url{http://nlp.stanford.edu/software/}}, including speech tagging, named-entity mentions, coreference resolution and dependency parsing. 

Moreover, we constructed a structured database from semi--structured HTML documents, the ``WWTTS--Maritime database'', containing 7,846 unique incidents, and utilised it as ground-truth data for distant supervision as discussed in Section~\ref{distant_supervision}. The WWTTS--Maritime database provides additional information regarding date and aggressor entities in each incident. We also used the ``WWTTS--Piracy database'', containing 7,895 unique incidents, which can be useful as an external database for the victim and incident type extractions (Section~\ref{distant_supervision}).

\subsection{The Ontology}
\vspace{-1.2mm}
A key step in the design of our knowledge graph is to define an ontology specifying which target relations to extract. We utilize the Simple Event Model (SEM) \cite{van2013simple}, which is a graph-based specification for events and related concepts, including actors, places and times. Inspired by the SEM ontology, we designed a relatively small maritime piracy event ontology with four main classes, \emph{Incident} (what happens), \emph{Actor} (who participated), \emph{Location} (where) and \emph{Date} (when).

The SEM ontology represents not only the description of who did what, when and where, but also the roles played by the \emph{Actors}. Each incident can have multiple actor types or incident types. Fig.~\ref{fig:SEM} shows how the SEM ontology can be applied to the analysis of maritime data. In this example, we focus on a container ship being attacked by pirates near position 05:28N –- 002:21E. In this maritime ontology, classes and sub-classes are nodes, while properties are edges of a graph. The \emph{Actor} has sub-classes including \emph{Individuals}, \emph{Organizations}, \emph{Transport ships}, \emph{Passenger ships}, \emph{Fishing ships}, \emph{Navy ships} and \emph{Other ships}. For instance, in Fig.~\ref{fig:SEM}, the ``container ship'' is a subclass of \emph{Transport ship}, which in turn, is a subclass of \emph{Actor}. 

In this work, we mainly define two roles for the extracted \emph{Actors}, which are {\it Victim} and {\it Aggressor}. Based on the context constraints, each \emph{Actor} has one of these roles.
In the next section, we explain how we map the data to this ontology. 

\subsection{Entity Extraction and Role Assignment}
\vspace{-1.2mm}
Maritime Deepdive seeks to extract different types of objects from input documents, namely \emph{entities}, \emph{entity mentions}, \emph{roles}, \emph{relations} and \emph{relation mentions}. An entity is a real-world place, actor or thing. In this work, \emph{Actor, Date, Location} and \emph{Incident--Type} are the entities of interest. An entity mention is a span of text in an input document that refers to an entity. We assign a role of \emph{Aggressor} or \emph{Victim} to any actor mention. For example, in Fig.~\ref{fig:SEM} ``container ship'' represents a mention for the actor entity which has the role of victim. A relation associates two or more entities, and represents the fact that there exists a relationship between the participating entities. A relation mention is a phrase that connects two entity mentions that participate in a relation. For instance, ``container ship'' and ``pirates'' participate in a Victim--Aggressor relation, which indicates that the container ship has been victimized by pirates. 

\begin{figure}[ht]
\vspace{-0.7em}
\scalebox{1}[1]{\includegraphics[width=1\linewidth]{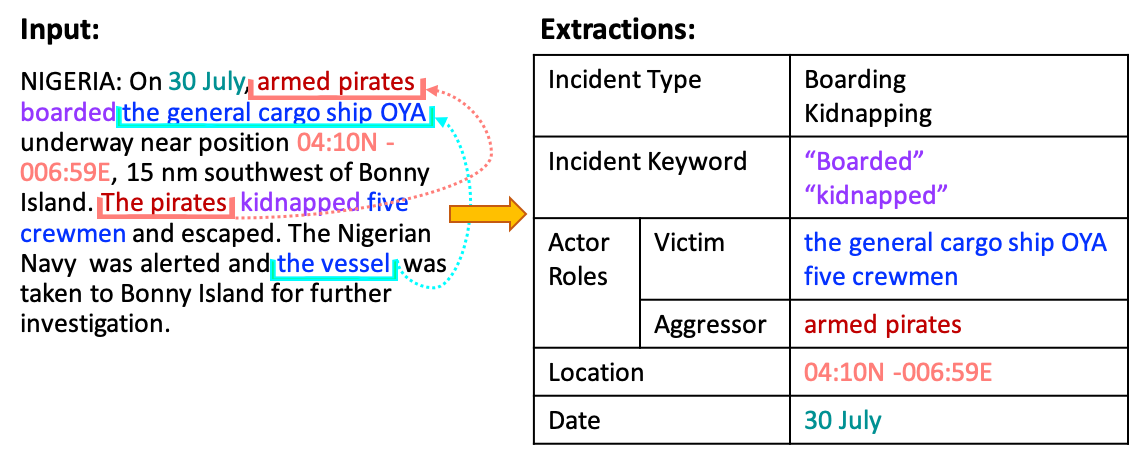}}
\caption{An illustration of entity extraction/linking and actor role assignments in Maritime DeepDive. The arrows denote coreferent mentions.}
\label{fig:extraction}
\vspace{-0.9em}
\end{figure}

\textbf{Entity Extraction:} Maritime DeepDive uses Stanford's CoreNLP natural language processing to extract useful markups and structure from the input data we provide. This step splits our incidents into sentences and their component tokens. Additionally, it returns lemmas, part-of-speech (POS) tags, named entity recognition (NER) tags, and a dependency parse of the sentence. For each incident, we extract a set of possible \emph{mentions} based on signals, e.g., positions, contained words, matched regular expressions and SQL queries.
For example, a simple User Define Function (UDF) written in Python or DDlog can tag spans of continuous tokens with the NER tag DATE as date entities. By involving a set of regular expressions we can extract the longitude and latitude of each incident as a location entity. For some of the incidents, there are no explicit coordinates of the location, but there is a textual description, such as in Example 1 below. For these incidents we either look up the coordinates of locations in the Google Maps geocoding API\footnote{https://developers.google.com/maps/documentation/geocoding/} or write a UDF to find the correct location entity mention using heuristic rules. We apply the same procedure to map all candidates for actor mentions and incident type entities, as seen in Fig.~\ref{fig:extraction}.

{\it{Example 1 -- BRAZIL: On 29 October, robbers boarded a passenger ferry on the Amazon River in the state of Para and  robbed the passengers and crew.}}


\begin{table*}[htbp]\renewcommand{\arraystretch}{1}
\caption{Some extracted features for a sample Victim--Date relation candidate.} 
\vspace{-1.0em}
\begin{center}
\setlength{\tabcolsep}{0.10em}
\renewcommand{\arraystretch}{0.98}
\begin{tabular}{|l|l|}
\hline
\multicolumn{2}{|l|}{REPUBLIC OF THE CONGO: On \textbf{7 January}, 12 seafarers, who were kidnapped from the Panama-flagged \textbf{tanker} Anuket Amber}\\ 
\multicolumn{2}{|l|}{and the Singapore-flagged anchor handling and supply vessel ARK TZE in October 2018 off the country’s coast, have been released}\\  
\multicolumn{2}{|l|}{and are all safe.}\\
\hline
\textbf{Description} & \textbf{Features} \\
\hline
The set of POS tags for the words between the mentions in the relation & \sffamily \tiny {$\begin{aligned} POS\_SEQ\_ [ \textrm{, CD  NNS , WP VBD VBN IN DT JJ}]\end{aligned}$} \\
\hline
The set of NER tags for the words between the mentions in the relation & \sffamily \tiny {$\begin{aligned} NER\_SEQ\_[\textrm{O NUMBER O O O O O O O MISC}]\end{aligned}$} \\
\hline
The set of lemmas of the words between the mentions in the relation &  \sffamily \tiny {$\begin{aligned} LEMMA\_SEQ\_[\textrm{, 12 seafarer , who be kidnap from the panama-flagged}] \end{aligned}$}\\
\hline
The set of words between the mentions in the relation &\sffamily \tiny {$\begin{aligned} WORD\_SEQ\_[\textrm{, 12 seafarers , who were kidnapped from the Panama-flagged}] \end{aligned}$ } \\
\hline
The sum of the lengths of the words in the mentions & \sffamily \tiny {$\begin{aligned} LENGTHS\_[1\_2]\end{aligned}$} \\
\hline
Indicator feature for whether the mentions start with a capital letter & \sffamily \tiny {$\begin{aligned} STARTS\_WITH\_CAPITAL\_[False\_False] \end{aligned}$}\\
\hline
The lemmas in a window of size 1 around the mentions &  
\sffamily \tiny {$W\_LEMMA\_L\_1\_R\_1\_[\textrm{on}]\_[\textrm{Anuket}]$}\\
\hline
The lemmas in a window of size 2 around the mentions & 
\sffamily \tiny {$W\_LEMMA\_L\_2\_R\_2\_[\textrm{: on}]\_[\textrm{Anuket Amber}] $} \\
\hline
The NERs in a window of size 1 around the mentions &  
\sffamily \tiny {$W\_NER\_L\_1\_R\_1\_[\textrm{O}]\_[\textrm{O}]$} \\
\hline
The NERs in a window of size 3 around the mentions &  
\sffamily \tiny  {$W\_NER\_L\_3\_R\_3\_[\textrm{LOCATION O O}]\_[\textrm{O O O}]$} \\
\hline
\end{tabular}
\label{features}
\end{center}
\vspace{-1.4em}
\end{table*}

\textbf{Entity Linking:} Entity linking is the process of mapping a textual mention to a real-world entity. We use coreference resolution with the CoreNLP coref annotator to link the entities\footnote{https://stanfordnlp.github.io/CoreNLP/coref.html}.
For example, in Fig.~\ref{fig:extraction}, two mentions of ``the general cargo ship OYA'' and ``the vessel'' refer to the same entity. Since the CoreNLP annotator mostly links noun entities, it is used for linking actor mentions. Entities can also be linked manually by adding heuristic rules. For instance, ``robbers'' and ``steal'' both indicate the incident type ``robbery'', and therefore consider the same entity for both mentions.


\textbf{Actor Role Assignment:} In this step, we assign roles to the actor mentions based on the incident context or distant supervision. Fig.~\ref{fig:extraction} indicates the roles assigned to different actor mentions. The rules are created by the user, and the process of creating these rules is based on the semantic context of the incident. For instance, in Fig.~\ref{fig:extraction}, the verb ``board'' appears immediately after the actor mentions (``armed pirates''). Hence, we will assign the {\it aggressor} role to the ``armed pirates'' mention.
Furthermore, the verb ``kidnap'' appears immediately before the actor mention (``five crewmen''). Therefore, we assign the {\it victim} role to this actor mention.  

\subsection{Relation candidate mapping and feature extraction}
\vspace{-1.2mm}
The next step within Maritime DeepDive is the extraction of relation mention candidates and features. For each binary relation, we take all pairs of non-overlapping entity mentions (e.g., victim and aggressor) that co-occur in an incident paragraph, and form Victim--Aggressor relation mention candidates. We declare eight binary relations, namely Victim--Aggressor, Victim--Date, Victim--IncidentType, Victim--Location, IncidentType--Date, Aggressor--IncidentType, Aggressor--Location and Aggressor--Date, and extract all the corresponding relation mention candidates. 

We also need to extract features associated with relation candidates. Maritime DeepDive includes an automatic feature generation library, DDlib. One baseline feature that is often used in DDlib is the word sequence between mention pairs in an incident. Table~\ref{features} shows a set of generic features extracted from Example 2 for a Victim--Date relation candidate.

{\it{Example 2 -- REPUBLIC OF THE CONGO: On \emph{7 January}, 12 seafarers, who were kidnapped from the Panama-flagged \emph{ tanker} Anuket Amber and the Singapore-flagged anchor handling and supply vessel ARK TZE in October 2018 off the  country’s coast, have been released and are all safe.}}

\subsection{Distant supervision}
\label{distant_supervision}
One challenge with building a machine learning system for knowledge graph construction is that it generally requires a large amount of labeled training data. However annotated data for maritime piracy incidents is not available, and manually labelling textual data is tedious and expensive. 
With Maritime DeepDive, we take the approach referred to as distant supervision, where we generate a noisy set of labels for candidate relation mentions using a mix of mappings from secondary datasets and heuristic rules. We implement distant supervision in two separate levels, entity-level and relation-level. We illustrate these procedures using examples.

\textbf{Distant supervision with secondary data:}
We utilise the WWTTS--Piracy and WWTTS--Maritime databases by assuming that they represent ``ground truth''. Some true and some false victim, aggressor, date and incident type entity mentions are generated by comparing the mention text to the ground truth data. 
For instance, Fig.~\ref{fig:extraction} describes an incident from our training dataset where its date, latitude and longitude match one or more incidents from the WWTTS--Piracy database. If the actor mention ``the general cargo ship OYA'' matches the ship type in the WWTTS--Piracy database, then we consider this mention as a true candidate for victim role in this incident. Moreover, in similar fashion, if the date, location and first 50 characters of this incident match an incident in the WWTTS--Maritime database, and the actor mention candidate ``armed pirates'', match the aggressor in the WWTTS--Maritime database, then we consider this mention as a true candidate for the aggressor role. These are examples of entity-level distant supervision. If in an incident both victim and aggressor candidates are labeled as true, then the binary relation of Victim--Aggressor is also labeled as true. Moreover, if any of victim or aggressor mentions in an incident are labeled as false, the binary relation between them are labeled as false. For instance, the relation mention ``the general cargo ship OYA-- armed pirates'' is a true candidate and the relation mention ``the general cargo ship OYA -- five crewmen'' is a false candidate for the Victim--Aggressor relation.

\textbf{Distant supervision with rules:}
We create supervision rules that do not rely on any secondary structured dataset, but instead just use some heuristic rules over the entities and relations. For instance in the entity level, based on the knowledge domain, we create a list of aggressor keywords such as \emph{robber, intruder, pirate, terrorist, gunman, assailant} and when an actor mention matches one of these keywords, the actor is very likely the aggressor. Furthermore, we create a list of aggressive acts such as \emph{attack, fire upon, hijack, board, kidnap}. Some incidents contain more than one date mention, as shown in Example 2. In this case we wrote a heuristic rule (UDF) to label the true date of the incident (for instance, the closest date to the victim mention is likely to be the true date).

We implemented this heuristic rule for the Victim--Aggressor relation: If an aggressive act comes via lemmas between the aggressor and victim mentions, the relation between them is marked as true. It can be observed in Fig.~\ref{fig:extraction} that the aggressive verb, ``board'', appears between ``armed pirates'' and ``the general cargo ship OYA''. Technically, ``board'' returns an identifier that determines a heuristically true label for the given relation mention. 

The distant supervision generates noisy, imperfect labels for some candidates. Machine learning techniques are able to exploit redundancy to cope with this noise and learn the relevant phrases (e.g., attack, board and kidnap). Negative examples are also generated by distant supervision to construct a balanced training set.

Finally, we describe a simple majority-vote approach to resolving multiple labels per example, which can be implemented within DDlog. These final labels train a machine learning classification model (here, as true or false mentions of a relation). A fraction (one tenth) of these labeled data is considered as a development dataset for the analysis and evaluation purpose in Section \ref{evaluation}.

\begin{figure*}[ht]
\vspace{-1.2mm}
\scalebox{1}[0.93]{\includegraphics[width=1\linewidth]{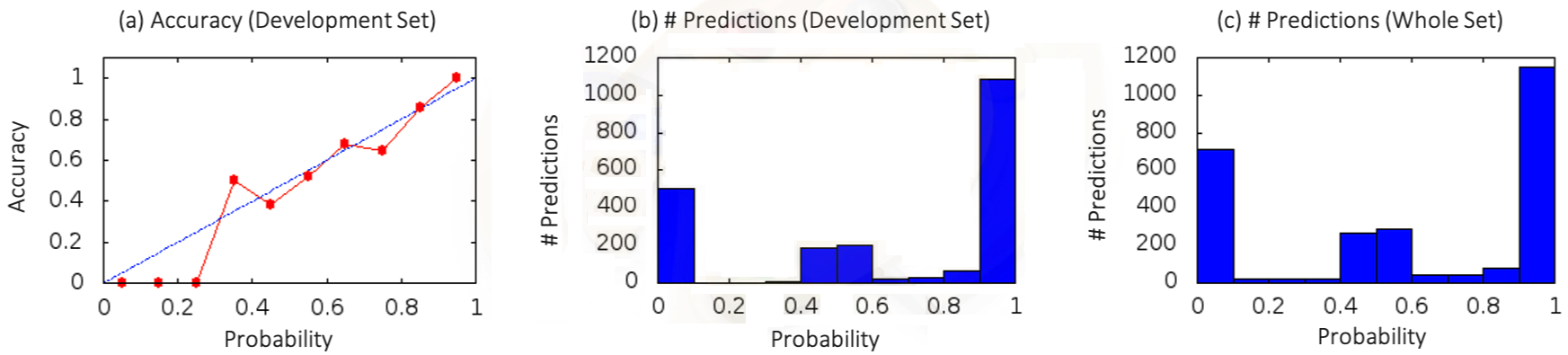}}
\vspace{-1.3em}
\caption{A sample calibration plot for the Aggressor--Location relation.}
\label{cal_plt}
\vspace{-1.4em}
\end{figure*}

\subsection{Learning and Inference}
Maritime DeepDive generates a factor graph that describes a set of random variables and how they are correlated. 
It can learn the weights of the factor graph from training data that can be either obtained through distant supervision or specified by the user while populating the database during the extraction phase. Maritime DeepDive performs statistical inference on the factor graph using standard techniques, such as Gibbs sampling \cite{wick2010scalable, zhang2013towards}. 
At the end of learning and inference, Maritime DeepDive outputs a marginal probability for  every candidate in the output knowledge base. As shown in Fig.~\ref{fig:pipeline}, the probabilistic labels generated by Maritime DeepDive for the Victim--Aggressor relation candidates are 0.94 (\textit{container ship}--\textit{pirates})  and 0.11 (\textit{Oil tanker}--\textit{crewman}).

Maritime Deepdive learns the parameters of the model (the weights of the features and potentially the connections between variables), and then performs statistical inference over the test dataset to estimate the probability that each variable of interest is true.\vspace{-3pt}

\begin{table}[htbp]\renewcommand{\arraystretch}{1}
\caption{Evaluation of our knowledge graph with respect to binary relation extraction performance. }
\vspace{-1.0em}
\begin{center}
\setlength{\tabcolsep}{0.10em}
\renewcommand{\arraystretch}{0.98}
\begin{tabular}{l|c|c|c|c|c}
\hline
\textbf{Relation} & \textbf{Candidates} &\textbf{ F1$ $}& \textbf{ROC-AUC} & \textbf{Recall} & \textbf{Precision}\\
\hline
Victim--Date & 142 & 0.96 & 0.97 & 0.96 & 0.96\\
\hline
Victim--IncidentType & 233 & 0.73 & 0.88 & 0.75 & 0.72\\
\hline
Victim--Aggressor & 196 & 0.72 & 0.82 & 0.75 & 0.80\\
\hline
Victim--Location & 154 & 0.90 & 0.88 & 0.92 & 0.88\\
\hline
IncidentType--Date & 114 & 0.85 & 0.75 & 0.86 & 0.85\\
\hline
Aggressor--IncidentType & 175 & 0.74 & 0.74 & 0.73 & 0.74\\
\hline
Aggressor–-Location & 115 & 0.72 & 0.67 & 0.81 & 0.67\\
\hline
Aggressor-–Date & 100 & 0.82 & 0.87 & 0.79 & 0.85\\
\hline
\end{tabular}
\label{f1}
\end{center}
\vspace{-1.2em}
\end{table}

\begin{table}[htbp]\renewcommand{\arraystretch}{1}
\caption{Evaluation of our knowledge graph on the binary relation extraction with distant supervision by only secondary databases, by only heuristic rules and with both distant supervision methods.}
\vspace{-1.0em}
\begin{center}
\begin{tabular}{l|c|c|c}
\hline
\textbf{Relation} &\multicolumn{3}{|c}{\textbf{F1-Score}}\\
\cline{2-4}
&Second Data &Heuristic  & Heu+Data\\
\hline
Victim--Date & 0.51 & 0.76  & \textbf{0.96} \\
\hline
Victim--IncidentType & 0.43 & 0.51  & \textbf{0.73} \\
\hline
Victim--Aggressor & 0.42 & 0.63  & \textbf{0.72}\\
\hline
Victim--Location & 0.12 & 0.75  & \textbf{0.90} \\
\hline
IncidentType--Date & 0.11 & 0.69  & \textbf{0.85} \\
\hline
Aggressor--IncidentType & 0.22 & 0.67  & \textbf{0.74} \\
\hline
Aggressor--Location & 0.44 & 0.71  & \textbf{0.72}\\
\hline
Aggressor--Date & 0.52 & 0.79  & \textbf{0.82} \\
\hline
\end{tabular}
\label{distant}
\end{center}
\vspace{-1.3em}
\end{table}

\section{Experimental evaluation}
\vspace{-1.2mm}
\label{evaluation}
In this section, we evaluate the performance of Maritime DeepDive. In many real-world (Defence) scenarios, annotated text data is scarce. Relying on human experts or crowdsourcing to manually annotate a large amount of text data is expensive or not always possible. 
As such, our major evaluation focus is on investigating the feasibility of automatically extracting events and activities of interest from maritime piracy reports  \textit{without any} annotated data (which is a challenging task). We organise this section into two stages. First, we carry out an evaluation on the quality (e.g., effectiveness and accuracy) of the constructed knowledge graph, where we measure F1, recall, precision and Receiver Operating Characteristics--Area Under the Curve (ROC--AUC) scores associated with the extracted binary relations. Next, we describe our process for the analysis and improvement of Maritime DeepDive's performance using \emph{calibration plots} without annotated data.

\begin{figure}[htbp]
\vspace{-1.0em}
\centering
{\label{fig:roc}\scalebox{1}[1]{\includegraphics[width=0.8\linewidth]{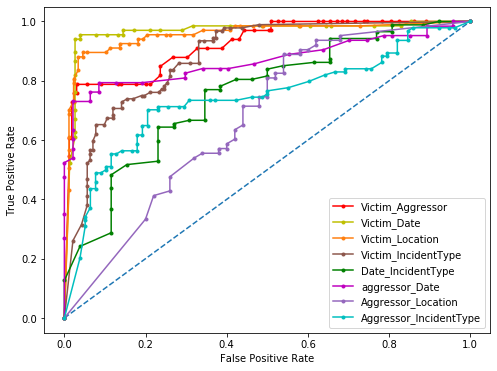}}}\\
{\label{fig:PR}\scalebox{1}[1]{\includegraphics[width=0.8\linewidth]{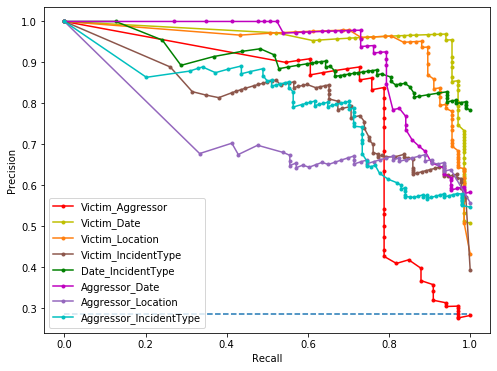}}}
\vspace{-1.3em}
\caption{Receiver Operating Characteristic (ROC) and precision–recall curves for the binary relations extracted using Maritime Deepdive. }
\label{fig:AUC_PR}
\vspace{-1em}
\end{figure}

\textbf{Relation extraction evaluation.}
For the evaluation of our model, we also focus on relation extraction within maritime data. By dynamically setting a threshold for probabilistic labels of each binary relation (i.e., only if the probability of a relation candidate is higher than the threshold can it be considered true), we can measure the F1 score, ROC--AUC, the curves of the ROC and the precision--recall of the output for each relation, as shown in Fig.~\ref{fig:AUC_PR}. 
We can see that in general, Maritime Deepdive is very effective at classifying all binary relations being investigated. Although there are certain relations where the classification results demonstrate excellent performance (e.g. {\it Victim--Date}), there are others where the classification results are good but not exceptional (e.g {\it Aggressor--Location}). In our experience, low precision occurs when Maritime DeepDive outputs a high probability for a fact, but it is an incorrect one or is not supported by the corresponding incident. We expect that this issue can be satisfactorily addressed by providing sufficient negative examples for training, or adding more distant supervision rules. Low recall occurs when Maritime DeepDive fails to extract a fact. This usually means that a relation candidate is not assigned a probability that is sufficiently high by Maritime DeepDive. In this case, one solution is to add more distant supervision rules to increase the number of positive examples.

We apply a relation--specific threshold to the generated probabilistic labels for all relation candidates. The threshold is set via maximizing the classification accuracy on the validation dataset. 
As mentioned in Section~\ref{data}, our test dataset contains 75 incidents.
The number of candidates and the predictive performance for each binary relation of Maritime DeepDive are listed in Table~\ref{f1}. 
Considering that we have a large unlabeled corpus, we are able to achieve satisfactory performance (the averages are 0.81 F1, 0.82 ROC-AUC, 0.82 recall and 0.81 precision scores) in the relation extraction task.

In addition, an important question is the significance of distant supervision on the predictive performance of Maritime DeepDive, where we only consider the effect of secondary databases as ground-truth for distant supervision. 
We compare the results of Maritime DeepDive with a simpler pipeline, which is trained via distant supervision using only  secondary databases (the second column of Table~\ref{distant}). Alternatively, we also trained Maritime Deepdive with heuristic rules for distant supervision (the third column of Table~\ref{distant}).
The last column of Table~\ref{distant} shows that the probabilistic labels of Maritime DeepDive with distant supervision from both heuristic rules and secondary databases 
consistently performs better than with either. We observe a significant improvement of $45\%$ on average over distant supervision with only secondary databases.
These results demonstrate that Maritime DeepDive more effectively learns from the heuristic and secondary datasets' probabilistic training labels over the simpler pipeline without heuristic or secondary datasets.
Given that we did not use any annotated (labelled) text as training data, the overall results obtained in Tables~\ref{f1} and~\ref{distant} are very promising.  We expect these results to be significantly enhanced once we add labelled training data and/or improve the labels generated by distant supervision using MindBender\footnote{Mindbender is an interactive search interface to browse data produced by DeepDive. \url{https://github.com/HazyResearch/mindbender}} and MindTagger\footnote{Mindtagger is a graphical user interface tool for labeling data. \url{http://deepdive.stanford.edu/labeling}}.



\textbf{Iterative improvement.}
Users can develop an intuitive sense of how to improve Maritime DeepDive from the high level summary of model performance provided by the calibration plots. Ideally, in Maritime DeepDive, the probability associated with a given fact should equal the empirical probability that this fact is correct, i.e., an extraction with a probability 0.95 should be correct 95$\%$ of the time when inspected in the original source. 

A calibration plot contains three components as a function of probability: (a) Accuracy, which measures the development set accuracy of a prediction; (b) $\#$ Predictions (Development set), which measures the number of extractions in the  development set with a certain probability; and (c) $\#$ Predictions (Whole Set), which measures the number of extractions in the whole set with a certain probability. 
For example, Fig.~\ref{cal_plt}.a shows the accuracy for the target relation Aggressor--Location (red line). In a well-tuned Maritime DeepDive system, this would coincide with the blue (0,0)-(1,1) line. The deviation from the (0,0)-(1,1) line can be caused by noise in the training data, quantization error, or sparsity in the training data. The histograms in Fig.~\ref{cal_plt}.b-c show satisfying results with U-shaped curves, with the majority of extractions concentrated at high probability and low probability and a few being indeterminate. In general, the goal of improving the accuracy of knowledge graph construction is to ``push'' these probabilities closer to either (0, 0.1) or (0.9, 1.0). The masses around 0.4-0.6 suggest that there are some types of candidates for which Maritime DeepDive has insufficient features or labels, resulting in low confidence/high uncertainty. Such results indicate relations for which additional distant supervision rules need to be added to resolve this uncertainty.\vspace{-5pt}


\section{Conclusion}
\vspace{-1.2mm}
We have  presented our initial approach to the automated construction of probabilistic knowledge graphs for the maritime domain, with a use case illustration for maritime piracy events. As many types of maritime incidents are recorded in a textual format (e.g., post-incident reports) and other associated activities such as pirates' communications and planning are entirely or partially conducted through words, being able to extract and construct a knowledge base from such textual data would not only provide  an effective method for query answering, but also facilitate further analysis and forecasting (e.g., identifying dangerous zones, spatial-temporal patterns, predicting trends and potentially discovering plans for  attacks). More significantly, the probabilistic knowledge graph constructed from text data can be combined with those constructed from other sources of data (e.g., rich contextual/domain knowledge and movement data generated by sensors as in \cite{nguyen2020fuzzy}) in order to provide comprehensive situational awareness and assist with the detection of illegal maritime activities. 


In future work, we plan to incorporate deep learning methods within our framework to increase robustness and generalisation on unseen data. Moreover, our proposed approach can be applied to threat detection in other domains, which we plan to investigate in the near future. \vspace{-5pt}
\bibliographystyle{IEEEtran}
\bibliography{IEEEabrv}

\begin{thebibliography}{10}
\providecommand{\url}[1]{#1}
\csname url@samestyle\endcsname
\providecommand{\newblock}{\relax}
\providecommand{\bibinfo}[2]{#2}
\providecommand{\BIBentrySTDinterwordspacing}{\spaceskip=0pt\relax}
\providecommand{\BIBentryALTinterwordstretchfactor}{4}
\providecommand{\BIBentryALTinterwordspacing}{\spaceskip=\fontdimen2\font plus
\BIBentryALTinterwordstretchfactor\fontdimen3\font minus
  \fontdimen4\font\relax}
\providecommand{\BIBforeignlanguage}[2]{{%
\expandafter\ifx\csname l@#1\endcsname\relax
\typeout{** WARNING: IEEEtran.bst: No hyphenation pattern has been}%
\typeout{** loaded for the language `#1'. Using the pattern for}%
\typeout{** the default language instead.}%
\else
\language=\csname l@#1\endcsname
\fi
#2}}
\providecommand{\BIBdecl}{\relax}
\BIBdecl

\bibitem{rosenberg2009political}
D.~Rosenberg, ``The political economy of piracy in the south china sea,''
  \emph{Naval War College Review}, vol.~62, no.~3, pp. 43--58, 2009.

\bibitem{hurlburt2013human}
K.~Hurlburt, ``The human cost of somali piracy,'' \emph{Piracy at Sea}, pp.
  289--310, 2013.

\bibitem{dabrowski2014contextual}
J.~J. Dabrowski, ``Contextual behavioural modelling and classification of
  vessels in a maritime piracy situation,'' Ph.D. dissertation, University of
  Pretoria, 2014.

\bibitem{rogova2013contextual}
G.~Rogova and J.~Garcia, ``{Contextual knowledge and information fusion for
  maritime piracy surveillance},'' \emph{Prediction and Recognition of Piracy
  Efforts Using Collaborative Human-Centric Information Systems}, vol. 109,
  2013.

\bibitem{rein2018beyond}
K.~Rein and J.~Biermann, ``Beyond situation awareness: Considerations for
  sense-making in complex intelligence operations,'' in \emph{2018 21st
  International Conference on Information Fusion (FUSION)}.\hskip 1em plus
  0.5em minus 0.4em\relax IEEE, 2018.

\bibitem{dragos2014integration}
V.~Dragos and K.~Rein, ``{Integration of soft data for information fusion:
  Pitfalls, challenges and trends},'' in \emph{17th International Conference on
  Information Fusion (FUSION)}.\hskip 1em plus 0.5em minus 0.4em\relax IEEE,
  2014.

\bibitem{wickramarathne2011belief}
T.~L. Wickramarathne, K.~Premaratne, M.~N. Murthi, M.~Scheutz, S.~K{\"u}bler,
  and M.~Pravia, ``{Belief theoretic methods for soft and hard data fusion},''
  in \emph{2011 International Conference on Acoustics, Speech and Signal
  Processing (ICASSP)}.\hskip 1em plus 0.5em minus 0.4em\relax IEEE, 2011.

\bibitem{nunez2013hard}
R.~C. N{\'u}nez, B.~Samarakoon, K.~Premaratne, and M.~N. Murthi, ``{Hard and
  soft data fusion for joint tracking and classification/intent-detection},''
  in \emph{Proceedings of the 16th International Conference on Information
  Fusion}.\hskip 1em plus 0.5em minus 0.4em\relax IEEE, 2013.

\bibitem{dragos2015critical}
V.~Dragos, X.~Lerouvreur, and S.~Gatepaille, ``{A critical assessment of two
  methods for heterogeneous information fusion},'' in \emph{2015 18th
  International Conference on Information Fusion (FUSION)}.\hskip 1em plus
  0.5em minus 0.4em\relax IEEE, 2015, pp. 42--49.

\bibitem{plachkov2016soft}
A.~Plachkov, ``{Soft Data-Augmented Risk Assessment and Automated Course of
  Action Generation for Maritime Situational Awareness},'' Ph.D. dissertation,
  Universit{\'e} d'Ottawa/University of Ottawa, 2016.

\bibitem{wu2014semantic}
K.~Wu, W.~Tang, K.~Mao, G.-W. Ng, and L.~O. Mak, ``{Semantic-level fusion of
  heterogenous sensor network and other sources based on Bayesian network},''
  in \emph{17th International Conference on Information Fusion (FUSION)}.\hskip
  1em plus 0.5em minus 0.4em\relax IEEE, 2014.

\bibitem{panchapakesan2019optimizing}
A.~Panchapakesan, R.~Abielmona, and E.~Petriu, ``{Optimizing Maritime Vessel
  Service Time with Adaptive Quay Crane Deployment Through Level 4 Hard-Soft
  Information Fusion},'' in \emph{2019 22th International Conference on
  Information Fusion (FUSION)}.\hskip 1em plus 0.5em minus 0.4em\relax IEEE,
  2019.

\bibitem{Abirami2015}
T.~{Abirami}, E.~{Taghavi}, R.~{Tharmarasa}, T.~{Kirubarajan}, and
  A.~{Boury-Brisset}, ``{Fusing social network data with hard data},'' in
  \emph{2015 18th International Conference on Information Fusion (FUSION)},
  2015.

\bibitem{burks2019collaborative}
L.~Burks and N.~Ahmed, ``{Collaborative semantic data fusion with dynamically
  observable decision processes},'' in \emph{2019 22th International Conference
  on Information Fusion (FUSION)}.\hskip 1em plus 0.5em minus 0.4em\relax IEEE,
  2019.

\bibitem{Buffett2017arctic}
S.~Buffett, C.~Cherry, C.~Dai, A.~Désilets, H.~Guo, D.~McDonald, J.~Su, and
  D.~Tulpan, ``{Arctic Maritime Awareness for Safety and Security (AMASS).
  Final Report},'' \emph{Canadian National Research Council}, 2017.

\bibitem{shapiro2015use}
S.~C. Shapiro and D.~R. Schlegel, ``{Use of background knowledge in natural
  language understanding for information fusion},'' in \emph{2015 18th
  International Conference on Information Fusion (FUSION)}.\hskip 1em plus
  0.5em minus 0.4em\relax IEEE, 2015.

\bibitem{zhang2020construction}
Q.~Zhang, Y.~Q. Wen, D.~Han, F.~Zhang, and C.~S. Xiao, ``{Construction of
  knowledge graph of maritime dangerous goods based on IMDG code},'' \emph{The
  Journal of Engineering}, vol. 2020, 2020.

\bibitem{nguyen2019combining}
V.~Nguyen, ``{On combining probabilistic and semantic similarity-based methods
  toward off-domain reasoning for situational awareness},'' in \emph{2019 22th
  International Conference on Information Fusion (FUSION)}.\hskip 1em plus
  0.5em minus 0.4em\relax IEEE, 2019.

\bibitem{nguyen2020fuzzy}
V.~Nguyen and L.~Mellor, ``{Fuzzy MLNs and QSTAGs for Activity Recognition and
  Modelling with RUSH},'' in \emph{2020 IEEE 23rd International Conference on
  Information Fusion (FUSION)}.\hskip 1em plus 0.5em minus 0.4em\relax IEEE,
  2020, pp. 1--8.

\bibitem{pelikanova2014google}
Z.~Pelik{\'a}nov{\'a}, ``{Google knowledge graph},'' 2014.

\bibitem{bollacker2008freebase}
K.~Bollacker, C.~Evans, P.~Paritosh, T.~Sturge, and J.~Taylor, ``{Freebase: a
  collaboratively created graph database for structuring human knowledge},'' in
  \emph{Proceedings of the 2008 ACM SIGMOD international conference on
  Management of data}, 2008.

\bibitem{rebele2016yago}
T.~Rebele, F.~Suchanek, J.~Hoffart, J.~Biega, E.~Kuzey, and G.~Weikum, ``{YAGO:
  A multilingual knowledge base from wikipedia, wordnet, and geonames},'' in
  \emph{International semantic web conference}.\hskip 1em plus 0.5em minus
  0.4em\relax Springer, 2016.

\bibitem{lehmann2015dbpedia}
J.~Lehmann, R.~Isele, M.~Jakob, A.~Jentzsch, D.~Kontokostas, P.~N. Mendes,
  S.~Hellmann, M.~Morsey, P.~Van~Kleef, S.~Auer \emph{et~al.}, ``{DBpedia--a
  large-scale, multilingual knowledge base extracted from Wikipedia},''
  \emph{Semantic web}, vol.~6, 2015.

\bibitem{ernst2015knowlife}
P.~Ernst, A.~Siu, and G.~Weikum, ``{Knowlife: a versatile approach for
  constructing a large knowledge graph for biomedical sciences},'' \emph{BMC
  bioinformatics}, vol.~16, 2015.

\bibitem{rotmensch2017learning}
M.~Rotmensch, Y.~Halpern, A.~Tlimat, S.~Horng, and D.~Sontag, ``{Learning a
  health knowledge graph from electronic medical records},'' \emph{Scientific
  reports}, vol.~7, 2017.

\bibitem{farber2019microsoft}
M.~F{\"a}rber, ``{The microsoft academic knowledge graph: a linked data source
  with 8 billion triples of scholarly data},'' in \emph{International Semantic
  Web Conference}.\hskip 1em plus 0.5em minus 0.4em\relax Springer, 2019.

\bibitem{chen2018knowedu}
P.~Chen, Y.~Lu, V.~W. Zheng, X.~Chen, and B.~Yang, ``{KnowEdu: a system to
  construct knowledge graph for education},'' \emph{IEEE Access}, vol.~6, 2018.

\bibitem{zhang2019construction}
Q.~Zhang, Y.~Wen, C.~Zhou, H.~Long, D.~Han, F.~Zhang, and C.~Xiao,
  ``{Construction of knowledge graphs for maritime dangerous goods},''
  \emph{Sustainability}, vol.~11, 2019.

\bibitem{everwyn2019link}
J.~Everwyn, A.-I. Mouaddib, B.~Zanuttini, S.~Gatepaille, and S.~Brunessaux,
  ``{Link Prediction on Dynamic Attributed Knowledge Graphs for Maritime
  Situational Awareness},'' in \emph{Conf{\'e}rence Nationale sur les
  Applications Pratiques de l’Intelligence Artificielle (APIA 2019)}, 2019.

\bibitem{de2016deepdive}
C.~De~Sa, A.~Ratner, C.~R{\'e}, J.~Shin, F.~Wang, S.~Wu, and C.~Zhang,
  ``{Deepdive: Declarative knowledge base construction},'' \emph{ACM SIGMOD
  Record}, vol.~45, 2016.

\bibitem{niu2011tuffy}
F.~Niu, C.~R{\'e}, A.~Doan, and J.~Shavlik, ``{Tuffy: Scaling up Statistical
  Inference in {Markov Logic Networks} using an RDBMS},'' \emph{Proceedings of
  the VLDB Endowment}, vol.~4, 2011.

\bibitem{van2013simple}
W.~R. van Hage and D.~Ceolin, ``{The simple event model},'' in \emph{Situation
  awareness with systems of systems}.\hskip 1em plus 0.5em minus 0.4em\relax
  Springer, 2013, pp. 149--169.

\bibitem{wick2010scalable}
M.~Wick, A.~McCallum, and G.~Miklau, ``{Scalable probabilistic databases with
  factor graphs and MCMC},'' \emph{arXiv preprint arXiv:1005.1934}, 2010.

\bibitem{zhang2013towards}
C.~Zhang and C.~R{\'e}, ``{Towards high-throughput Gibbs sampling at scale: A
  study across storage managers},'' in \emph{Proceedings of the 2013 ACM SIGMOD
  International Conference on Management of Data}, 2013.

\end{thebibliography}

\end{document}